\documentclass[letterpaper, 10 pt, conference]{ieeetran} 
\IEEEoverridecommandlockouts
\newcommand\copyrighttext{%
	\footnotesize This work has been submitted to the IEEE for possible publication. Copyright may be transferred without notice, after which this version may no longer be accessible.
}
\newcommand\copyrightnotice{%
	\begin{tikzpicture}[remember picture,overlay]
	\node[anchor=south,yshift=10pt, xshift=10pt] at (current page.south) {\fbox{\parbox{\dimexpr\textwidth-\fboxsep-\fboxrule\relax}{\copyrighttext}}};
	\end{tikzpicture}%
}

\usepackage{printlen}

\usepackage[T1]{fontenc}
\usepackage{amsmath,amsfonts, amssymb}
\usepackage{cite}
\usepackage{stfloats}
\usepackage{graphicx}
\usepackage[caption=false,font=scriptsize]{subfig}
\usepackage{url}
\usepackage{hyperref}

\usepackage[per-mode = fraction]{siunitx}
\usepackage{acronym}
\usepackage[none]{hyphenat}
\usepackage{glossaries}

\usepackage{bbm}

\usepackage{algorithm}
\usepackage{algpseudocode}
\usepackage[frozencache,cachedir=minted-cache]{minted}

\usepackage{booktabs}
\usepackage{tabularx}
\usepackage{multirow}
\usepackage{pifont}

\usepackage{tikz, pgfplots, pgfplotstable}
\usetikzlibrary{arrows, arrows.meta, patterns}
\usepgfplotslibrary{statistics}
\pgfplotsset{compat=1.18}
\usetikzlibrary{external}
\tikzset{external/only named=true}

\usepackage{xcolor}
\definecolor{Black}{HTML}{000000}
\definecolor{Blue}{HTML}{0065bd}
\definecolor{Bluelight}{HTML}{D6E8F7}
\definecolor{Bluestrong}{HTML}{003359}
\definecolor{Red}{HTML}{8C000F}
\definecolor{Orange}{HTML}{E37222}
\definecolor{OrangePP}{HTML}{E97132}
\definecolor{Green}{HTML}{A2AD00}
\definecolor{GreenCR}{HTML}{008000}
\definecolor{LightGray}{HTML}{e7e7e7}
\definecolor{Gray}{HTML}{7f7f7f}
\definecolor{Gray-opac}{HTML}{d8d8d8}
\definecolor{MyDarkBlue}{RGB}{14,40,65}

\definecolor{TUMBlueBrand}{HTML}{3070b3}
\definecolor{TUMBlueDark}{HTML}{072140}
\definecolor{TUMBlueDark5}{HTML}{165db1}
\definecolor{TUMBlueLight}{HTML}{5e94d4}
\definecolor{TUMBlueBright}{HTML}{8f81ea}
\definecolor{TUMOrange}{HTML}{f7b11e}
\definecolor{TUMRed}{HTML}{ea7237}

\graphicspath{{./figures/}}

\title{\LARGE \bf
Modular Autonomy with Conversational Interaction: An LLM-driven Framework for Decision Making in Autonomous Driving
}

\newif\iffinal
\finaltrue

\newcommand{\authorsFinal}{
    \author{Marvin Seegert, Korbinian Moller, Johannes Betz%
    \thanks{M.Seegert, K. Moller, and J. Betz are with the Professorship of Autonomous Vehicle Systems, TUM School of Engineering and Design, Technical University of Munich, 85748 Garching, Germany; Munich Institute of Robotics and Machine Intelligence (MIRMI).}%
    }
}

\newcommand{\authorsBlind}{%
  \author{Anonymous Author(s)%
  \thanks{This work has been submitted to IV 2026 for peer review. Affiliations are omitted for double-blind review.}%
  }
}

\iffinal
  \authorsFinal
\else
  \authorsBlind
\fi

\newacronym{ads}{ADS}{Autonomous Driving System}%
\newacronym{av}{AV}{Autonomous Vehicle}%
\newacronym{llm}{LLM}{Large Language Model}%
\newacronym{vlm}{VLM}{Vision Language Model}%
\newacronym{ros2}{ROS2}{Robot Operating System 2}%
\newacronym{dsl}{DSL}{Domain Specific Language}%
\newacronym{icl}{ICL}{in-context learning}%

\begin{document}
\bstctlcite{BSTcontrol}

\maketitle
\copyrightnotice

%%%%%%%%%%%%%%%%%%%%%%%%%%%%%%%%%%%%%%%%%%%%%%%%%%%%%%%%%
%%% Abstract
%%%%%%%%%%%%%%%%%%%%%%%%%%%%%%%%%%%%%%%%%%%%%%%%%%%%%%%%%

\begin{abstract}
Recent advancements in \glspl{llm} offer new opportunites to create natural language interfaces for \glspl{ads}, moving beyond rigid inputs. This paper addresses the challenge of mapping the complexity of human language to the structured action space of modular \gls{ads} software. We propose a framework that integrates an \gls{llm}-based interaction layer with Autoware, a widely-used open-source software. This system enables passengers to issue high-level commands, from querying status information to modifying driving behavior. Our methodology is grounded in three key components: a taxonomization of interaction categories, an application-centric \gls{dsl} for command translation, and a safety-preserving validation layer. A two-stage \gls{llm} architecture ensures high transparency by providing feedback based on the definitive execution status. Evaluation confirms the system's timing efficiency and translation robustness. Simulation successfully validated command execution across all five interaction categories. This work provides a foundation for extensible, \gls{dsl}-assisted interaction in modular and safety-conscious autonomy stacks.

\end{abstract}

%%%%%%%%%%%%%%%%%%%%%%%%%%%%%%%%%%%%%%%%%%%%%%%%%%%%%%%%%
%%% Content
%%%%%%%%%%%%%%%%%%%%%%%%%%%%%%%%%%%%%%%%%%%%%%%%%%%%%%%%%

%%%%%%%%%%%%%%%%%%%%%%%%%%%%%%%%%%%%%%%%%%%%%%%%%%%%%%%%%
%%% Introduction
%%%%%%%%%%%%%%%%%%%%%%%%%%%%%%%%%%%%%%%%%%%%%%%%%%%%%%%%%
\section{Introduction}
\label{sec:introduction}

\acrfullpl{ads} have made significant progress in recent years, supported by modular software frameworks such as Autoware~\cite{Zhao2024}. While these systems offer robust perception, planning, and control, their interaction interfaces are typically reactive to the environment or just hardcoded inputs~\cite{Teng2023}. With the emergence of \acrfullpl{llm}, new possibilities arise to bridge human intent and machine autonomy through natural language. \glspl{llm} have shown remarkable generalization across tasks, including instruction following, question answering, and domain adaptation~\cite{Li2024-Survey}.
Recent efforts have begun exploring the integration of \glspl{llm} into \gls{ads} pipelines, for instance, enabling passengers to influence driving behavior or request explanations through natural language~\cite{Wang2023, Cui2024-DriveLLM, Song2025}. However, many approaches are restricted to simulation or simplistic real-world setups and lack integration with established frameworks. Furthermore, real-world evaluation under operational constraints such as system latency, message safety, and runtime behavior remains limited~\cite{Wang2023, Cui2024-DriveLLM, Song2025}.

In this paper, we present a framework for \gls{llm}-assisted co-pilot interaction with Autoware, enabling passengers to influence and query the \gls{ads} using natural language. Unlike end-to-end control architectures~\cite{Chen2024}, our approach focuses on high-level interaction: commands and questions are interpreted by an \gls{llm}, categorized by intent, and mapped to valid interactions with existing Autoware modules. The system supports multiple interaction types, while ensuring operational robustness, through operating entirely on top of Autoware’s existing safety architecture. 

We implement and evaluate the system in a simulation environment, enabling the direct transfer to our research vehicle for real-world driving scenarios. We test the system in different realistic simulation scenarios, analyzing robustness, response latency, and user command coverage. While limited in scope, e.g., it does not yet provide causality explanations for driving decisions, the system demonstrates the practical feasibility of combining modular \gls{ads} with \gls{llm} interfaces. Our open-source implementation lays the foundation for future extensions, including domain-adaptive reasoning, and multimodal interaction. In summary, this paper has the following contributions: 

\begin{figure}[!t] 
    \centering 
    \includegraphics[width=\linewidth]{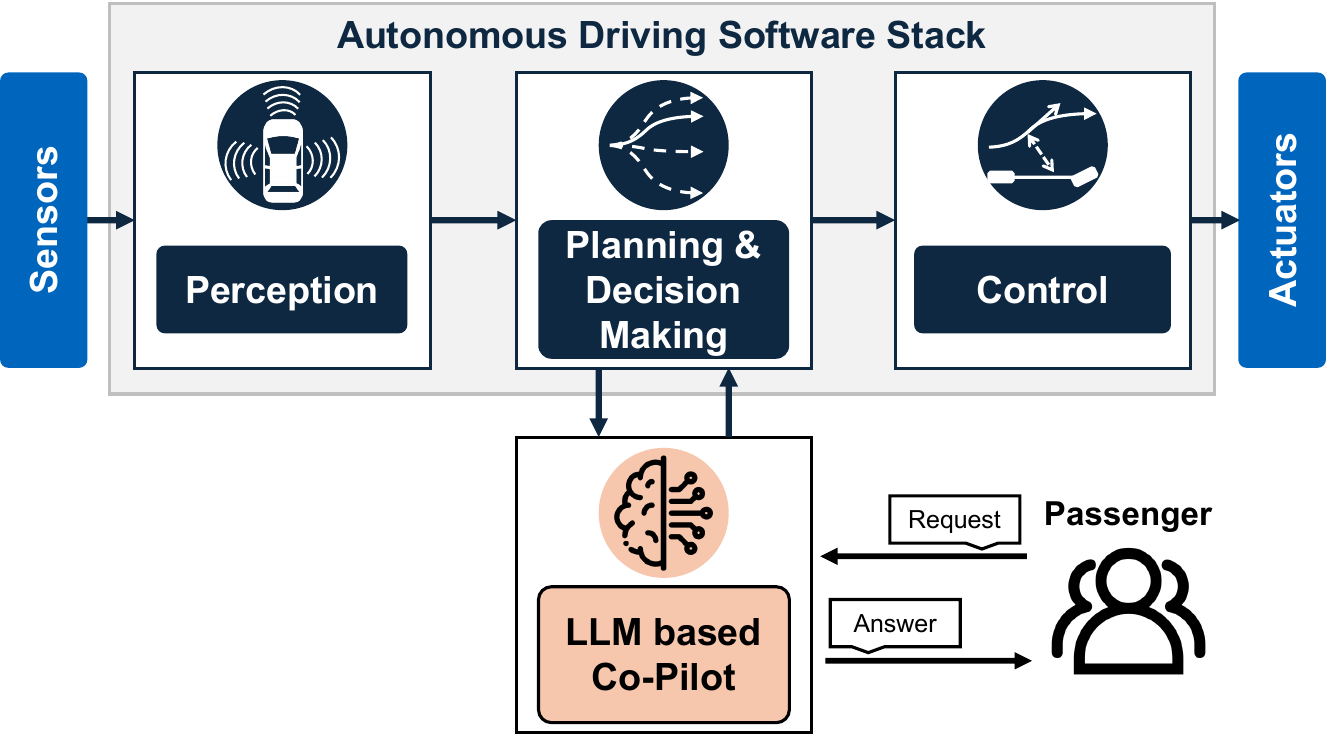} 
    \caption{Conceptual diagram of the \gls{llm}-based co-pilot interfacing natural language instructions and requests with the modular \gls{ads} software stack.} 
    \label{fig:intro} 
\end{figure}

\begin{enumerate}
    \item We present a framework that enables natural language interfaces for Autoware-based \gls{ads}. The system supports high-level command execution, integrated without modifying safety-relevant driving functions.
    \item We introduce a categorization of interaction types in autonomous driving, including representative examples, which lay the foundation for a framework-independent Domain Specific Language for Human-\gls{ads}-Interactions.
    \item We deploy and evaluate our framework in simulation, demonstrating feasibility under realistic driving conditions.
\end{enumerate}

%%%%%%%%%%%%%%%%%%%%%%%%%%%%%%%%%%%%%%%%%%%%%%%%%%%%%%%%%
%%% Related Work
%%%%%%%%%%%%%%%%%%%%%%%%%%%%%%%%%%%%%%%%%%%%%%%%%%%%%%%%%
\section{Related Work}
\label{sec:relatedwork}

Foundation models, such as \glspl{llm} and \glspl{vlm}, are increasingly investigated in the context of autonomous driving. They have been applied to tasks ranging from scene understanding to trajectory reasoning and high-level decision support. In the following, we first outline recent work on the general use of foundation models in autonomous driving. We then focus on hybrid approaches that couple natural language processing with structured \gls{ads} software stacks, enabling user interaction through language-based commands and queries.

\subsection{Foundation Models in Autonomous Driving}
Several studies have investigated the integration of \glspl{llm} and \glspl{vlm} into different stages of the autonomous driving pipeline. The authors of DriveGPT4~\cite{Xu2024, Xu2025}, LMDrive~\cite{Shao2024}, and SimLingo~\cite{Renz2025} employ multimodal transformer architectures that fuse camera or LiDAR inputs with linguistic features to generate low-level control signals or waypoints in an end-to-end fashion. These models demonstrate better interpretability and scene comprehension through natural language reasoning and achieve strong results in simulation. However, they remain constrained to simulated environments due to latency, data dependency, and the limited verifiability of large models in safety-critical control loops.

VLP~\cite{Pan2024} and DiMA~\cite{Hegde2025} use \glspl{llm} to inject semantic priors and common sense reasoning into vision-based planners, while DualAD~\cite{Wang2025-DualAD} combines textual reasoning with rule-based motion planning to enhance robustness in complex traffic situations. Similar works treat \glspl{llm} as high-level reasoning agents that guide behavior planning across multiple tasks~\cite{Azarafza2024, Liu2023, Fu2024}.

At the perception level, Li et al.~\cite{Li2024}, Wu et al.~\cite{Wu2025}, and Ran et al.~\cite{Ran2025} deploy language-grounded approaches for object tracking, segmentation, and scene explanation. By linking visual and textual representations, these works improve the semantic understanding and interpretability of driving scenes. Nevertheless, real-world validation remains limited, and the integration of foundation models into operational ADS frameworks has yet to be realized.

\subsection{Natural Language Interaction and Human-in-the-Loop Autonomous Driving}
Another line of work explores how \glspl{llm} can facilitate communication between humans and \glspl{av}, enabling explainable interaction, adaptive behavior, and shared decision-making.

Wang et al.~\cite{Wang2023} were among the first to propose an \gls{llm}-based co-pilot that interprets user intent and driving style preferences to adapt vehicle behavior, demonstrating the potential of conversational control, though limited to simulated scenarios.
Subsequent approaches extended this idea toward interactive reasoning. Cui et al.~\cite{Cui2024-DriveAsYouSpeak, Cui2024-DriveAsYouSay} treat the \gls{llm} as a cognitive core that interprets spoken instructions and explains its decisions, while the authors in~\cite{Cui2024-DriveLLM, Cui2024-FieldExperiments} realize real-world prototypes, integrating \gls{llm} reasoning and personalization into full software stacks. These systems support interaction and field-tested user adaptation, but are limited to a few specific driving scenarios.

Song et al.~\cite{Song2025} advance this concept with Autoware.Flex, translating natural language commands into modifications of Autoware modules, which makes it not directly applicable to other software frameworks, and tests remain limited to specific low-speed scenarios.
Other studies focus on user understanding rather than control. Luo et al.~\cite{Luo2025} introduce an interpreter that provides visual and spoken feedback to explain vehicle actions, thereby enhancing user trust. Choudhary et al.~\cite{Choudhary2024} present Talk2BEV, which allows querying BEV scenes through language, demonstrating explainable perception in simulation.

Beyond technical frameworks, Stampf et al.~\cite{Stampf2024} explore \gls{llm}-powered negotiation strategies between passengers and \glspl{av}, showing that adaptive conversational styles enhance trust and acceptance.
Overall, existing research successfully demonstrates that language interfaces can enhance transparency, personalization, and cooperation within ADS. However, most current systems either rely on simplified setups or directly delegate safety-critical decisions to the \gls{llm} itself. Consequently, a modular, safety-preserving framework that interfaces natural language requests with established \gls{ads} architectures, enabling a wide range of interactions, from configuration parameters to local scenario-based decisions, remains largely unexplored.

To address this gap, we introduce an application-centric framework that uses a dedicated validation layer. This approach safely maps high-level user commands to the vehicle's structured action space, decoupling language interpretation from the safety-critical execution loop.

%%%%%%%%%%%%%%%%%%%%%%%%%%%%%%%%%%%%%%%%%%%%%%%%%%%%%%%%%
%%% Methodology
%%%%%%%%%%%%%%%%%%%%%%%%%%%%%%%%%%%%%%%%%%%%%%%%%%%%%%%%%
\section{Methodology}
\label{sec:method}
The core challenge is translating spoken instructions~$\mathcal{I}$ from an infinite, continuous language space into a structured, executable policy~$\mathcal{P}$. This policy consists of finite actions~$\mathcal{A}$, but may include continuous parameters $\mathcal{C}$, making the target~$\mathcal{P} = \{\mathcal{A}, \mathcal{C}\}$ a continuous, bounded policy space. \glspl{llm} have been proven to be promising for this kind of complex high-level task, even without extensive retraining of the underlying model.

\begin{figure*}[!ht] 
    \centering 
    \includegraphics[width=\linewidth]{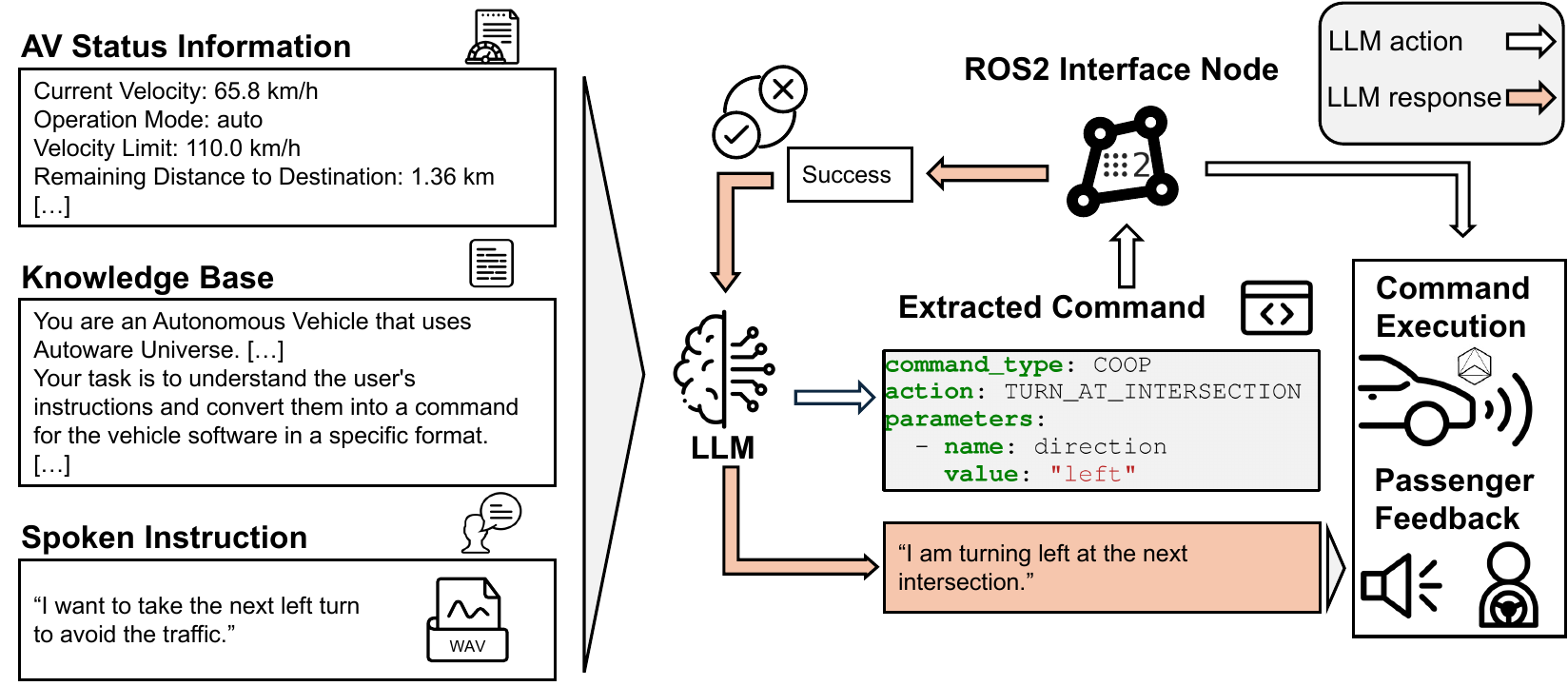}
    \vspace{-3mm}
    \caption{Illustration of the proposed \gls{llm}-based architecture for processing natural language spoken instructions into executable commands for an \gls{av}. The \gls{llm} integrates \gls{av} status information, a contextual knowledge base, which defines the \gls{llm}'s role and contains examples, and the Spoken Instruction. It then translates this input into a structured Extracted Command in a \gls{dsl} format. The Extracted Command is routed to a \gls{ros2} Interface Node for validation and subsequent Command Execution by the \gls{av}'s control systems. Successful execution triggers a confirmation, enabling appropriate Passenger Feedback.} 
    \label{fig:methodology_overview} 
\end{figure*}

Based on this, we propose the use of an \gls{llm} as the backbone of a general framework for the execution of spoken instructions by a passenger using a modular \gls{av} software stack, as shown in \autoref{fig:methodology_overview}. A spoken instruction~$\mathcal{I}$ together with status information from the \gls{av} software and a knowledge base, which consists of general information about the task as well as detailed descriptions about the expected output format, are forwarded to the \gls{llm}. The output is an extracted command, which is executed by the \gls{av} software using a validation and interface node. This framework is designed to bridge the gap between the infinite complexity of human language and the structured requirements of executing tasks with the \gls{av} software.

\subsection{Taxonomy of Human-ADS-Interactions}
As demonstrated in Section~\ref{sec:relatedwork}, existing approaches that enable human input using \glspl{llm} are often restricted to the execution of specific tasks. However, this paper aims to establish a general framework for connecting \glspl{llm} to a modular \gls{av} software stack, thus enabling the execution of a wide range of abstract human commands and requests. 

To address the challenge of mapping the infinite instruction space $\mathcal{I}$ to the structured, finite action set $\mathcal{A}$, we first establish a theoretical categorization of different use cases and tasks that a human passenger might seek to achieve by interacting with the \gls{av} system. This categorization serves two primary purposes: First, defining a system scope by clearly delineating the specific interaction types that can be facilitated by a language-based system and second, providing a necessary taxonomy for better describing and evaluating the capabilities of language-based interaction systems. This can not only be applied in the scope of this work but also for future research and standardization efforts in autonomous driving.

An overview of the five categories of passenger interaction with the \gls{av} software is shown in \autoref{tab:interaction_categories}. 
The \textit{Information} category covers requests for specific data, like vehicle velocity or software operation mode. The system queries and presents this data without executing a driving action. 
\textit{Mission Control} requests serve as direct input for the mission planning component, such as setting a destination or starting/stopping the autonomous drive. Enabling this via spoken language, rather than just operator input, increases system flexibility.
The \textit{Configuration} category includes requests for long-term behavioral changes, such as setting velocity or acceleration limits. These parameters allow passengers to adjust the overall driving experience to their individual comfort level.
\textit{Cooperation} focuses on requests for local driving maneuvers, like lane changes. The passenger initiates the request, but the \gls{av} software performs the final execution subject to its own safety checks, making this a high-level planning interaction.
The \textit{Intervention} category holds requests to override the \gls{av}'s internal state, typically due to false detections or decisions, such as a misread traffic light, ghost objects, or undetected objects. The passenger's command serves as supplementary information for the planning process, enabling a safe resolution of the situation.

\begin{table*}[!ht]
\centering
\caption{Classification of Passenger Interaction Categories for Autonomous Vehicle Systems}
\label{tab:interaction_categories}
\begin{tabularx}{\textwidth}{lXX}
\toprule
\textbf{Category} & \textbf{Description} & \textbf{Example Passenger Requests} \\
\toprule
\multirow{2}{*}{\textbf{Information}} & Requests \textbf{status data} about the vehicle's state or the \gls{av} software's internal condition for user clarity. & \begin{minipage}[t]{\hsize}\begin{itemize}
    \item What is the current speed limit?
    \item When do we arrive at the destination?
\end{itemize}\end{minipage} \\
\midrule
\multirow{2}{*}{\textbf{Mission Control}} & \multirow{2}{*}{Manages the \textbf{global operating state} and defines high-level goals.} & 
\begin{minipage}[t]{\hsize}\begin{itemize}
    \item Set destination to Campus Building.
    \item Start driving autonomously.
\end{itemize}\end{minipage} \\
\midrule
\multirow{2}{*}{\textbf{Configuration}} & Defines \textbf{long-term driving preferences} that adjust general behavioral parameters. & 
\begin{minipage}[t]{\hsize}\begin{itemize}
    \item Keep more distance to the car in front.
    \item Accelerate less aggressive.
\end{itemize}\end{minipage} \\
\midrule
\multirow{2}{*}{\textbf{Cooperation}} & Triggers \textbf{local driving maneuvers} in specific traffic scenarios. Execution is subject to safety checks by the \gls{av} software. & 
\begin{minipage}[t]{\hsize}\begin{itemize}
    \item Overtake the car in front of me on the highway.
    \item Turn right when it's possible.
\end{itemize}\end{minipage} \\
\midrule
\multirow{2}{*}{\textbf{Intervention}} & \textbf{Corrects errors} in perception or planning modules to override incorrect mandatory stops or system misinterpretations. & 
\begin{minipage}[t]{\hsize}\begin{itemize}
    \item The traffic light is green.
    \item Emergency, stop now.
\end{itemize}\end{minipage} \\
\bottomrule
\end{tabularx}
\end{table*}

\subsection{Language Interface for Passenger Interaction}
\begin{listing}[b]
\caption{Example command to set the maximum allowed velocity used in the planning algorithms to $v=90.0 \ \frac{\text{km}}{\text{h}}$.}
\label{lst:dsl}
\begin{minted}[
    frame=single, 
    framesep=5pt,
    bgcolor=gray!10,
    linenos=false,
    fontsize=\small
]{yaml}
command_type: CONFIG           (<CATEGORY>)
action: SET_PARAM                (<ACTION>)
parameters:
  - name: max_vel                  (<NAME>)
    value: 90.0                   (<VALUE>)
\end{minted}
\end{listing}

Based on the categorization of passenger interactions defined in \autoref{tab:interaction_categories}, a \acrfull{dsl} is proposed, which accommodates all defined actions belonging to these interaction categories. The primary task of the \gls{llm} is to translate the natural language user request~$\mathcal{I}$ into a command~$\mathcal{P}$, structured in this specific \gls{dsl} format, making it interpretable and executable by the \gls{av} software stack.

In contrast to existing \glspl{dsl}, such as AutoIR~\cite{Song2025}, our proposed \gls{dsl} is intentionally not structured around the specific architecture of the used software stack and its modules. Instead, it is structured based on the application-centric categories and actions requested. This categorization-based approach offers robustness against internal structural changes in the \gls{av} software and a clearer separation of concerns between user intent and system implementation. Due to this, our \gls{dsl} can be used in conjunction with Autoware, as well as other modular \gls{av} software stacks.

An example of our proposed \gls{dsl} is shown in Listing~\ref{lst:dsl}. It consists of the action $a \in \mathcal{A}$ and its interaction category, as well as name-value pairs for action specific parameters $\mathcal{C}_a$, which may be necessary depending on the action. The execution of each action is handled by a Validation and Interface node.

\subsection{Integration into Autoware Stack}
\begin{figure}[!b] 
    \centering 
    \includegraphics[width=0.98\linewidth]{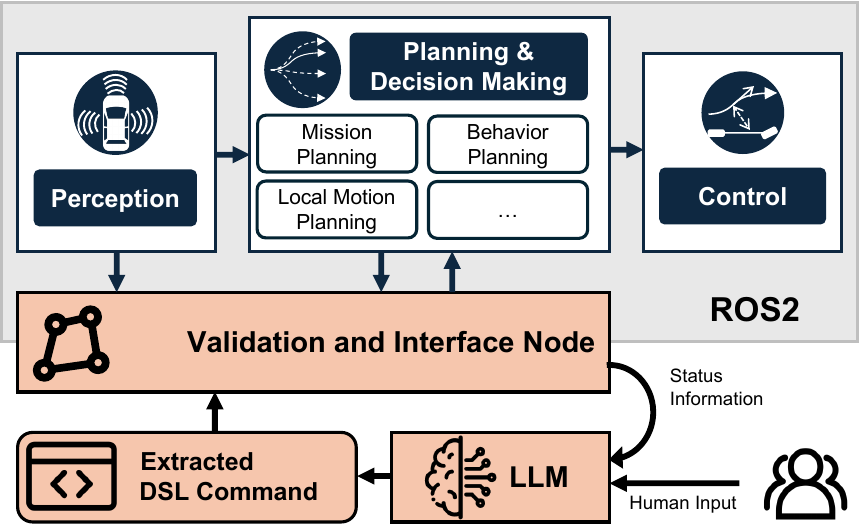} 
    \vspace{-3mm}
    \caption{Integration architecture featuring the Validation and Interface Node, which connects the extracted \gls{dsl} command from the \gls{llm} to the modular \gls{av} software stack, especially the Planning and Decision Making module, via \gls{ros2}.} 
    \label{fig:integration} 
\end{figure}

To enable the execution of the \gls{dsl} commands within Autoware, a dedicated \gls{ros2} Interface Node is employed. This node is responsible for parsing the structured information from the \gls{dsl} command and triggering the corresponding execution by sending the appropriate \gls{ros2} command, service call, or parameter adjustment. The entire integration architecture is conceptually illustrated in \autoref{fig:integration}.

The interface node also acts as an essential Action Validation Layer. The mapping from a \gls{dsl} action to a specific execution in Autoware must be built manually and explicitly. This explicit mapping acts as a safety measure, ensuring that only allowed actions with validated parameter bounds are executed. Without this verification, the system could be compromised by false interpretations, \gls{llm} hallucinations, or malicious commands, posing a serious safety risk. This explicit mapping, therefore, maintains the separation of concerns.

\subsection{Two-Stage Architecture for Feedback Generation}
To avoid on-board GPU constraints, we use a cloud-based Gemini~2.5~Flash-Lite \gls{llm}, which also allows us to bypass a separate Speech-to-Text (STT) step by processing audio instructions directly.
For high system transparency and reliable passenger feedback, a two-stage generation architecture is used, outlined in \autoref{fig:methodology_overview}. This approach utilizes a feedback loop to provide a response based on the actual outcome of the requested action.

The process begins in the first stage, where the \gls{llm} translates the spoken instruction into the structured \gls{dsl} command. This command is immediately forwarded to the \gls{ros2} Interface Node for validation and execution within the Autoware stack.

The second stage commences after the Interface Node receives the definitive success or failure state from the Autoware stack. Crucially, this execution status, along with the original user intent, is fed back into the \gls{llm} for a second API call. This two-step process ensures that the final passenger feedback is accurate and contextually relevant, reflecting the actual outcome of the action. This approach also provides essential system transparency to the user by communicating the rationale for any rejected command.

The \gls{llm} generated response is then converted into an audio file using a Text-to-Speech (TTS) framework, which is played back to the passenger as audio feedback.

%%%%%%%%%%%%%%%%%%%%%%%%%%%%%%%%%%%%%%%%%%%%%%%%%%%%%%%%%
%%% Results
%%%%%%%%%%%%%%%%%%%%%%%%%%%%%%%%%%%%%%%%%%%%%%%%%%%%%%%%%
\section{Results \& Discussion}
\label{sec:results}
The evaluation of the proposed system includes quantitative results evaluating the translation from human instruction to the corresponding command using a custom dataset and qualitative results, demonstrating the basic functionality in a simulation environment. Both evaluations were executed on a desktop computer with an Intel\textregistered\ Core\texttrademark\ i5-14500 CPU, which is also running the Autoware planning simulation.

\subsection{Action Translation Experiments}
For the quantitative evaluation of our architecture, a dataset of $N=200$ pairs of human instructions, along with their corresponding \gls{dsl} commands is used. This dataset was created manually and includes instructions from all five interaction categories with a roughly equal distribution, as well as an additional category, which includes all requests, that are outside the scope of this system. The dataset consists of partly repeating \gls{dsl} commands, but the instructions vary by their tone and the directness.

This dedicated instruction dataset is utilized to assess the translation quality and response time of our system. Human instructions from this set are first synthesized into an audio file using a TTS model to simulate human input. These spoken instructions are then forwarded to the \gls{llm}. Our full \textit{Baseline} system receives three primary inputs: status information from the \gls{av} software, the system's knowledge base (KB), which outlines the task and the \gls{dsl} structure, and a small set of \gls{icl} examples. To evaluate the contribution of these inputs, we defined three ablation configurations: \textit{No AV Status} (removes dynamic vehicle context), \textit{Zero-Shot} (removes \gls{icl} examples), and \textit{KB-Only} (removes both \gls{av} status and \gls{icl}). This evaluation is focused on the instruction translation step, comparing the \gls{llm}'s predicted \gls{dsl} output to the ground truth.

The results of this ablation study are presented in \autoref{tab:ablation_study}. The \textit{Baseline} system demonstrated high performance, correctly translating $194$ instructions for an accuracy of $97.0$\%. Removing the \gls{av} status information had a negligible impact, resulting in $96.0$\% accuracy, as most dataset instructions were not context-dependent. In contrast, removing the \gls{icl} examples caused a significant performance collapse. The \textit{Zero-Shot} accuracy dropped to $73.0$\%, and the \textit{KB-Only} system fell further to $66.3$\%. This indicates that while the \gls{llm} can perform tasks using only the \gls{dsl} definition, the \gls{icl} examples are critical for achieving robust and reliable translation accuracy. 

\begin{table}[ht]
\centering
\caption{Ablation study on the $N=200$ instruction dataset, with all experiments including the knowledge base (KB). Accuracy denotes successful translation to the ground-truth \gls{dsl} command, $\overline{t_r}$ and $\tilde{t_r}$ are the mean and median translation times.}
\label{tab:ablation_study}
\begin{tabular}{lccccc}
\toprule
\textbf{Experiment} & \textbf{\gls{av} Status} & \textbf{\acrshort{icl}} & \textbf{Acc. in \%} & \textbf{$\overline{t_r}$ in s} & \textbf{$\tilde{t_r}$ in s} \\
\midrule
\textit{Baseline} & \checkmark & \checkmark & $97.0$ & $1.723$ & $1.669 $\\
\textit{No \gls{av} Status} & $\times$ & \checkmark & $96.0$ & $1.735$ & $1.674$ \\
\textit{Zero-Shot} & \checkmark & $\times$ & $73.0$ & $1.703$ & $1.683$ \\
\textit{KB-Only} & $\times$ & $\times$ & $66.3$ & $1.717$ & $1.689$ \\
\bottomrule
\end{tabular}
\end{table}

An analysis of the six translation errors in our \textit{Baseline} run revealed that only one mistake occurred at the parameter level (confusing lateral and longitudinal acceleration). The remaining five were specification errors at the category and action level, showing interpretation discrepancies by the \gls{llm}.

\begin{figure}[ht]
    \centering
    \begin{tikzpicture}
\begin{axis}[
    boxplot/draw direction=x,
    ylabel={},
    xlabel={Translation times $t_r$ in s},
    width=10cm,
    height=6cm,
    xmin=1.22,
    xmax=4.8,
    yticklabels={},
    ytick= ,
    xtick align=outside,
    xtick pos=bottom,
    ytick style={draw=none},
    legend style={
        at={(0.98,0.98)},
        anchor=north east,
        draw=black,
        fill=white,
        rounded corners=1pt,
        line width=0.6pt,
        inner sep=2pt,
        /tikz/every even column/.append style={column sep=5pt}
    },
    legend image code/.code={%
        \draw[thick,#1] (0cm,0cm) -- (0.3cm,0cm);
    },
]

\addplot[TUMBlueBrand, thick] coordinates {(0,0)}; 
\addlegendentry{\textit{Baseline}}
\addplot[TUMBlueDark, thick] coordinates {(0,0)}; 
\addlegendentry{\textit{No \gls{av} Status}}
\addplot[TUMOrange, thick] coordinates {(0,0)}; 
\addlegendentry{\textit{Zero-Shot}}
\addplot[TUMBlueBright, thick] coordinates {(0,0)}; 
\addlegendentry{\textit{KB-Only}}

\addplot[TUMRed, thick] coordinates {(0,0)}; 
\addlegendentry{Song et al. \cite{Song2025}}

% --- Autoware Flex data
\addplot+[
    color=TUMRed,
    solid,
    boxplot prepared={
        lower whisker=1.65,
        lower quartile=2.56,
        median=2.96,
        upper quartile=3.40,
        upper whisker=4.64
    },
] coordinates {};
\addplot+[only marks, mark=x, color=TUMRed] coordinates {(4.689,1)};

% --- Add abl0 to abl3 datasets

    \addplot+[
        color=TUMBlueBright,
        solid,
        boxplot prepared={{
            lower whisker=1.35,
            lower quartile=1.58,
            median=1.669,
            upper quartile=1.79,
            upper whisker=2.10
        }},
    ] coordinates {};
    \addplot+[only marks, mark=x, color=TUMBlueBright] coordinates {(2.13,2) (2.16,2) (2.89,2) (3.15,2) (3.16,2) (3.19,2) (3.31,2)};

    \addplot+[
        color=TUMOrange,
        solid,
        boxplot prepared={{
            lower whisker=1.32,
            lower quartile=1.56,
            median=1.674,
            upper quartile=1.84,
            upper whisker=2.26
        }},
    ] coordinates {};
    \addplot+[only marks, mark=x, color=TUMOrange] coordinates {(2.37,3) (3.27,3)};

    \addplot+[
        color=TUMBlueDark,
        solid,
        boxplot prepared={{
            lower whisker=1.36,
            lower quartile=1.57,
            median=1.683,
            upper quartile=1.81,
            upper whisker=2.17
        }},
    ] coordinates {};
    \addplot+[only marks, mark=x, color=TUMBlueDark] coordinates {(2.19,4) (2.25,4) (2.29,4) (2.35,4) (2.38,4) (3.32,4) (3.35,4) (3.46,4)};

    \addplot+[
        color=TUMBlueBrand,
        solid,
        boxplot prepared={{
            lower whisker=1.32,
            lower quartile=1.57,
            median=1.689,
            upper quartile=1.82,
            upper whisker=2.21
        }},
    ] coordinates {};
    \addplot+[only marks, mark=x, color=TUMBlueBrand] coordinates {(2.25,5) (2.30,5) (2.31,5) (2.33,5) (2.56,5) (3.36,5)};

\end{axis}
\end{tikzpicture}
    \vspace{-6mm}
    \caption{Boxplot comparison of system response times $t_r$ for the \gls{llm} translation task. The results from the $N=200$ dataset for our proposed system (\textcolor{TUMBlueBrand}{\rule[0.5ex]{0.2cm}{1pt}}) are compared against the same system with ablations and against the results from Song et al.~\cite{Song2025}~(\textcolor{TUMRed}{\rule[0.5ex]{0.2cm}{1pt}}), which used a different dataset of same size and a similar, but non-identical experimental setup with a different \gls{llm} model and \gls{dsl}. Note on data extraction: The characteristic values for the boxplot representing~\cite{Song2025} were derived using a digital plot-digitizer tool from the published figure, as the original numerical data were unavailable.}
    \label{fig:boxplots}
\end{figure}
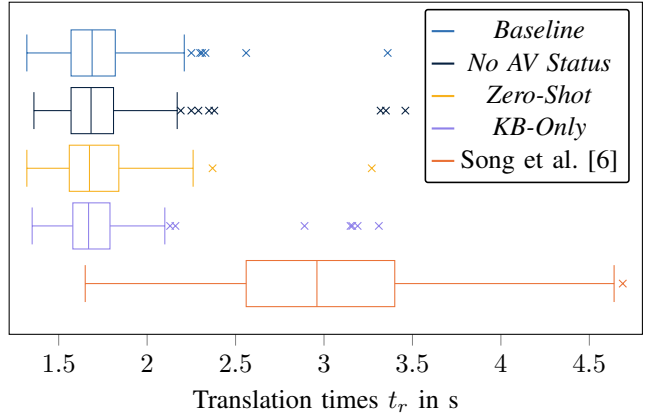

The response time $t_r$ for all four configurations was measured, with results visualized in \autoref{fig:boxplots}. As the boxplots for all our configurations clearly illustrate, the ablations had no significant impact on latency, suggesting that the inference time is dominated by the core generation task, with negligible overhead from processing \gls{icl} examples or \gls{av} status.

When compared to related work, our \textit{Baseline} system demonstrates significant advantages over Song~et~al.~\cite{Song2025}. Our accuracy of $97.0$\% exceeds their overall translation accuracy of $87.0$\%. Furthermore, as illustrated in \autoref{fig:boxplots}, our \textit{Baseline} latency is significantly lower than their reported results. The observed jitter is also notably smaller for our system, indicating a more consistent and predictable user experience. We hypothesize this performance difference in both speed and consistency is due to the application-centric design of our proposed system. By abstracting commands, our \gls{dsl} simplifies the translation task for the \gls{llm}, compared to the more complex module-centric approach in \cite{Song2025}. The specific choice of the \gls{llm} and its API may also play a role.

\subsection{Evaluation in Simulation}
\begin{figure*}[!h] 
    \centering 
    \includegraphics[width=\linewidth]{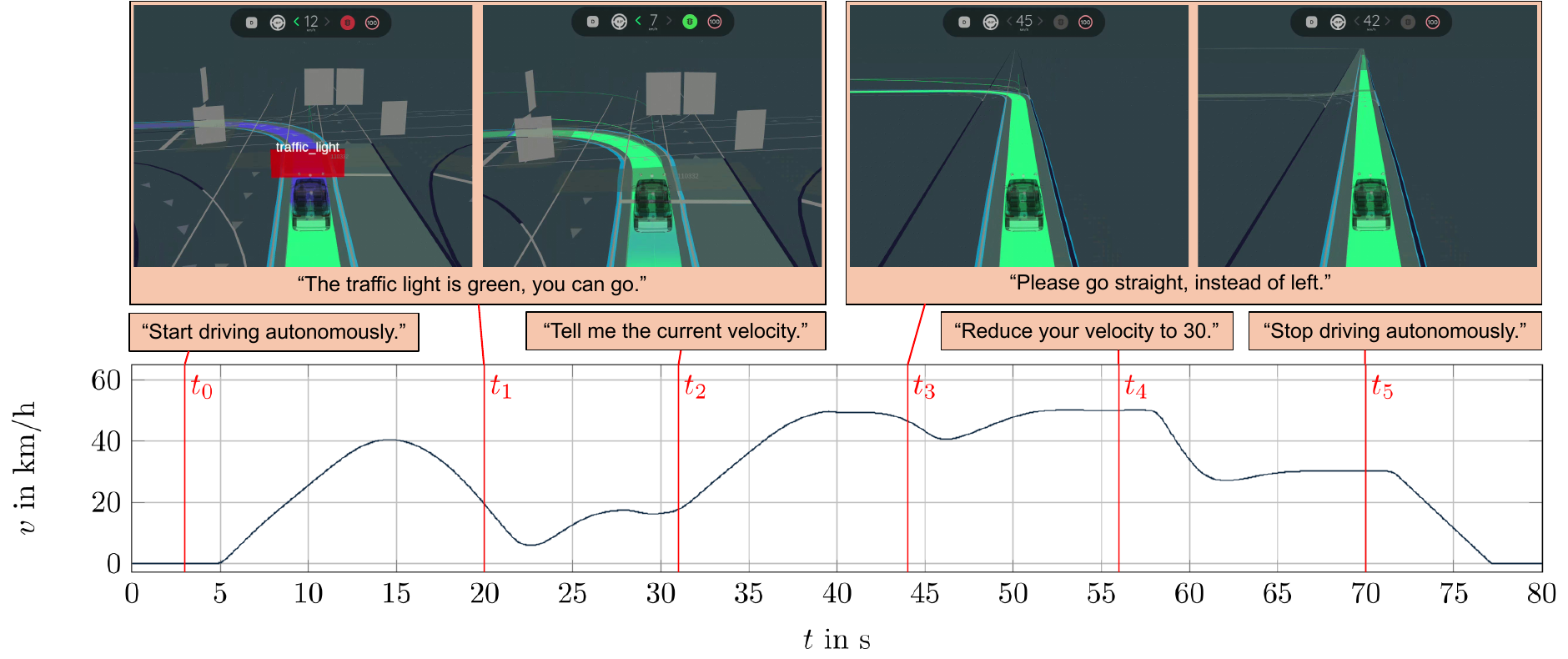} 
    \vspace{-6mm}
    \caption{Time-series plot illustrating the vehicle's velocity profile $v$ over time $t$ in the Autoware planning simulation environment, showcasing the execution of multiple passenger instructions. At $t_0$ - $t_5$ spoken instructions were issued by the system and executed by the \gls{av} software.} 
\label{fig:simres} 
\end{figure*}

The system's full integration was first confirmed by successfully deploying the framework on our real-world research vehicle, where it executed basic commands. Due to vehicle availability, a comprehensive quantitative evaluation of all five interaction categories was conducted in a simulation environment. Given that the proposed system primarily interacts with the planning and decision making modules and does not rely on direct sensor data usage, the Autoware planning simulation was chosen as the testbed.

\begin{figure}[t] 
    \centering 
    \includegraphics[width=\linewidth]{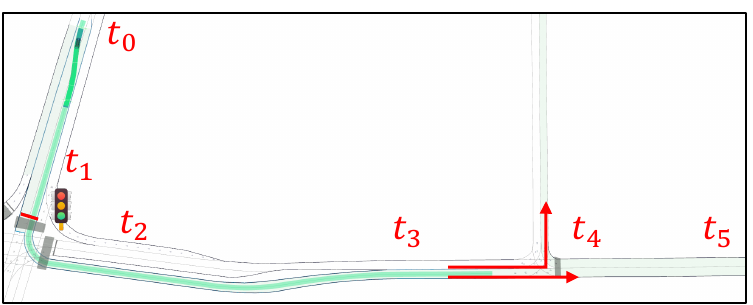} 
    \vspace{-6mm}
    \caption{The driven route for in the Autoware planning simulation. At $t_1$ is a traffic light, that is overridden by the user instruction and at $t_3$ the planned route is going to turn left at the intersection, but the instruction is requesting to go straight instead.} 
\label{fig:map} 
\end{figure}

A defined test scenario was created as illustrated in \autoref{fig:simres}, to validate the system's handling of all five interaction categories. This scenario was executed on a custom vector map based on a real-world street topology shown in \autoref{fig:map}. The scenario includes:
\begin{itemize}
    \item \textit{Mission Control} (starting/stopping the drive at $t_0, t_5$)
    \item \textit{Intervention} (overriding a simulated red light at $t_1$)
    \item \textit{Information} (querying vehicle status at $t_2$)
    \item \textit{Cooperation} (re-planning a route at an intersection at $t_3$)
    \item \textit{Configuration} (setting a new speed limit at $t_4$)
\end{itemize}

\noindent
To quantitatively assess the full system reliability and performance, this complete scenario was executed $N=30$ times. The aggregated results are summarized in \autoref{tab:simulation_results}.

\begin{table}[htb]
\centering
\caption{Quantitative reliability and performance results from $N=30$ simulation runs. Latencies are reported as mean~$\overline{t}$, median~$\tilde{t}$, and standard deviation~$\sigma$.}
\label{tab:simulation_results}
\begin{tabular}{lc}
\toprule
\textbf{Performance Metric} & \textbf{Value} \\
\midrule
\multicolumn{2}{l}{\textbf{System Reliability}} \\
\ Successful Runs $N$ & $29$/$30$ \\
\ Task Success Rate $TSR$ & $96.7$\%  \\
\ Failure Source & 1x API Server Error \\
\midrule
\multicolumn{2}{l}{\textbf{Component Latencies}} \\
\ Command Execution (in ms) & $\overline{t}=5.737$, $\tilde{t}=0.461$, $\sigma=8.491$ \\
\ Feedback Generation (in s) & $\overline{t}=1.846$, $\tilde{t}=1.778$, $\sigma=0.355$ \\
\bottomrule
\end{tabular}
\end{table}

The system demonstrated high reliability, achieving a Task Success Rate (TSR) of $96.7$\%, with $29$ of the $30$ runs completing successfully. The single failure was traced to an API server error during an \gls{llm} request for feedback generation.

The component latencies from the $29$ successful runs are also detailed in \autoref{tab:simulation_results}. The Command Execution latency exhibits a highly skewed distribution, with a mean time of $\overline{t}=5.737$ ms, a high standard deviation of $\sigma=8.491$ ms, and a low median $\tilde{t}=0.461$ ms. This skew is attributed to the mix of near-instantaneous \gls{ros2} topic publications and slower service calls required for execution of certain actions. However, with a maximum latency in the millisecond range, this execution time is orders of magnitude smaller than the \gls{llm} inference times and is therefore considered negligible in the end-to-end latency budget.

The Feedback Generation latency, which represents the second \gls{llm} call, added a predictable median latency of~$\tilde{t}=1.778$ s. This result is consistent with the initial translation times reported in \autoref{tab:ablation_study}, confirming the predictable and stable overhead of the two-stage architecture.

Collectively, this $N=30$ experimental setup confirmed the system's applicability in a realistic, full-stack scenario using Autoware. It quantitatively demonstrated that the system translates and executes commands from all five interaction categories successfully with high reliability (96.7\% TSR) and features a computationally efficient validation loop.

%%%%%%%%%%%%%%%%%%%%%%%%%%%%%%%%%%%%%%%%%%%%%%%%%%%%%%%%%
%%% Conclusion
%%%%%%%%%%%%%%%%%%%%%%%%%%%%%%%%%%%%%%%%%%%%%%%%%%%%%%%%%
\section{Conclusion \& Outlook}
\label{sec:conclusion}
This paper introduced a \gls{llm}-based interaction framework designed to enable passengers to influence the behavior of an autonomous vehicle running a modular software stack. A key contribution is the categorization of five interaction types, which forms the basis for a robust, application-centric \gls{dsl}. Our translation evaluation demonstrated a high \textit{Baseline} accuracy of $97.0$\% at $\overline{t_r} = 1.723$~s. An ablation study confirmed that \gls{icl} was critical to this result, as the \textit{Zero-Shot} accuracy dropped significantly to $73.0$\%. The system's safety and transparency are ensured by a dedicated Validation and Interface Node and a two-stage generation architecture. This design was quantitatively validated in $N=30$ simulation runs, demonstrating high reliability with a~$96.7$\% Task Success Rate across all five interaction categories. We also confirmed the architecture's efficiency, with negligible command execution latencies. This two-stage design is vital, as it ensures passenger feedback reflects the definitive execution status, thereby building user trust.

Despite these promising results, this work has several limitations: the reliance on a cloud-based \gls{llm} API introduces network-dependent latency and a single point of failure, as seen in our reliability test. Furthermore, the \gls{llm} is not fine-tuned for the driving domain, and its robustness to ambiguity was not exhaustively tested. Finally, our quantitative evaluation was limited to a defined scenario and requires testing across more diverse traffic situations.

While acknowledging these limitations, our framework provides a solid foundation for future extensions. Future work will focus on enhancing user trust by providing domain-adaptive explanations. Furthermore, the \gls{llm}'s confirmed ability to understand driving contexts lays the groundwork for advanced autonomous decision-making. By integrating an \gls{llm} or \gls{vlm} into the high-level decision loop, the system could potentially move towards the autonomous resolution of edge-case scenarios.

%%%%%%%%%%%%%%%%%%%%%%%%%%%%%%%%%%%%%%%%%%%%%%%%%%%%%%%%%
%%% Acknowledgement
%%%%%%%%%%%%%%%%%%%%%%%%%%%%%%%%%%%%%%%%%%%%%%%%%%%%%%%%%

\iffinal
\section*{Acknowledgement}
\noindent This project has been partly funded by the German Federal Ministry of Research, Technology and Space (BMFTR) within the project 'ASUR' under grant number 03ZU2105BA.
\fi

%%%%%%%%%%%%%%%%%%%%%%%%%%%%%%%%%%%%%%%%%%%%%%%%%%%%%%%%%%%%%%%%%%%%%%%%%%%%%%%%

%%%%%%%%%%%%%%%%%%%%%%%%%%%%%%%%%%%%%%%%%%%%%%%%%%%%%%%%%
%%% Bibliography
%%%%%%%%%%%%%%%%%%%%%%%%%%%%%%%%%%%%%%%%%%%%%%%%%%%%%%%%%
\bibliographystyle{IEEEtran}
%\clearpage
\bibliography{literature}
\end{document}